\documentclass[11pt,letterpaper]{article}
\usepackage{emnlp2016}
\usepackage{times}
\usepackage{latexsym}

\usepackage{url}
\usepackage{color}
\usepackage{float}
\usepackage{tabularx}
\usepackage{subfig}
\usepackage{multirow}
\usepackage{latexsym}
\usepackage{amsmath}
\usepackage{amssymb}
\usepackage{multirow}
\usepackage{pgfplots}
\usepackage{algorithm}
\usepackage{algorithmic}

\def\emnlppaperid{***}

\DeclareMathOperator*{\argmax}{arg\,max}

\def\h{\mathbf{h}}
\def\e{\mathbf{e}}

\def\p{\mathbf{p}}

\def\bb{\mathbf{b}}

\def\bW{\mathbf{W}}

\def\R{\mathbf{R}}

\allowdisplaybreaks

\newenvironment{itemize*}%
  {\begin{itemize}%
    \setlength{\itemsep}{0pt}%
    \setlength{\parskip}{0pt}}%
  {\end{itemize}}
  \newenvironment{enumerate*}%
  {\begin{enumerate}%
    \setlength{\itemsep}{0pt}%
    \setlength{\parskip}{0pt}}%
  {\end{enumerate}}
\emnlpfinalcopy
\title{Neural Sentence Ordering}

\date{}
\author{Xinchi Chen, Xipeng Qiu, Xuanjing Huang\\
 Shanghai Key Laboratory of Intelligent Information Processing, Fudan University\\
School of Computer Science, Fudan University\\
825 Zhangheng Road, Shanghai, China\\
\{xinchichen13, xpqiu, xjhuang\}@fudan.edu.cn}

\begin{document}

\maketitle

\begin{abstract}
Sentence ordering is a general and critical task for natural language generation applications. Previous works have focused on improving its performance in an external, downstream task, such as multi-document summarization. Given its importance, we propose to study it as an isolated task. We collect a large corpus of academic texts, and derive a data driven approach to learn pairwise ordering of sentences, and validate the efficacy with extensive experiments. Source codes\footnote{https://github.com/fudannlp} and dataset\footnote{http://nlp.fudan.edu.cn/data/} of this paper will be made publicly available.
\end{abstract}

\section{Introduction}

The goal of sentence ordering is to arrange a set of sentences into a coherent text in a clear and consistent manner \cite{grosz1995centering,van1999semantic,barzilay2008modeling}. The task is general and yet challenging, and is especially important for natural language generation \cite{reiter1997building}. Its applications include multi-document summarization, question answering, and concept-to-text generation. Improper ordering of sentences can generate confusing texts, degrading  readability.

A text should be organized according to it discourse coherence of the following properties: rhetorical \cite{mann1988rhetorical} coherence \cite{hobbs1990literature}, topical relevancy, chronological sequence, and cause-effect \cite{hume1750philosophical,okazaki2004improving}. These properties intertwine with each other, and can be quite subtle, as shown in the example in Table \ref{tab:example}.

\begin{table}[t] 
\centering
\begin{tabular}{|p{0.4\textwidth}|}
  \hline
  (1) He liked music when he was a boy.\\
  (2) People are shocked by his potential.\\
  (3) Chopin is a great musician in Poland.\\
  (4) When he was 15, he finished his first waltz.\\
  \hline
  Gold: (3) (1) (4) (2)\\ \hline
\end{tabular}
\caption{Illustration of sentence ordering task. Multiple discourse coherence relations might appear in a single text. First sentence (3) declares a topic. Sentences (1) (4) are in chronological sequence. However, sentence (2) is a result, so it should be the last sentence so as to abide to a cause-effect relation.}\label{tab:example}
\end{table}

Most of previous researches of sentence ordering were integrated into an external and downstream task, such as multi-document summarization \cite{barzilay2002inferring,lapata2003probabilistic,bollegala2010bottom}. The input sentences are extracted from multiple sources, therefore their intrinsic  coherence is relatively weak. Consequently, it is somewhat difficult to judge of the order of given sentences.
Moreover, these methods addressed the ordering problem of newspaper articles. Ordering criteria include majority ordering, chronological ordering, topical-closeness, precedence, and succession.  Among them, chronological ordering (i.e. orders sentence by the publication date) can produce satisfactory orderings \cite{barzilay2002inferring,okazaki2004improving}. Obviously, this is a natural result for ordering sentences extracted from newspaper articles, since the task is to arrange a large number of time-series events concerning several topics. Besides, all these criteria can be considered as reasonable hand-engineered features. Nevertheless, they cannot be adapted to other tasks or domains, as our example shows.


In this paper, we stage this problem as a standalone task, and adopt a data driven approach. We first derive neural model to encode each sentence into distributed representation (dense vector), then predict the pairwise ordering of sentences. Next, to avoid brute-force rearrangement, we use a beam search to determine the most probable permutation.

For this purpose, we collect about a million abstracts of research papers from arXiv website\footnote{https://arxiv.org/}. These abstracts are well designed coherent texts, and each of them involves several different criteria, including chronological ordering, topical-closeness, etc. For instance, abstracts might first declare the shortcomings of previous methods, leading to the reason why they propose the new one. That is cause-effect relation. Chronological sequences (marked by keywords ``first'', ``then'', etc.) might appear when they describe their models.

 The contributions of this paper can be summarized as follows:
 \begin{enumerate*}
   \item We frame sentence ordering as an isolated task, and collected a large corpus whose correct ordering goes beyond conventional criteria.
   \item Instead of relying on hand-designed features, we explore a fully data-driven approach to {\it learn} the order of a set of sentences.
   \item We perform extensive empirical studies and demonstrate the efficacy of our approach.
 \end{enumerate*}

\section{Sentence Ordering}
\subsection{Task Description}\label{sec:Task_Describe}
Sentence ordering task takes a text $s$ that is possibly out-of-order sentences,
\begin{equation}
s = s_1, s_2, \dots, s_{n_s}.
\end{equation}
and finds the \emph{gold} order. A good model must has the ability to capture the logic of a text.
That is,
the goal is to discover an order $o$, which is equal to the gold order $o^*$ of these sentences:
\begin{equation}
  s_{o^*_1} \succ s_{o^*_2}\succ \dots\succ s_{o^*_{n_s}},
\end{equation}

Here, $o$ is one of permutations of numbers in $\{1,2,\dots, n_s\}$. For instance, in Table \ref{tab:example}, whereas the current order $o$ is $[1, 2, 3, 4]$, $o = o* = [3, 1, 4, 2]$ is the gold order.

\subsection{Ranking Model} \label{sec:Model}
\begin{figure}[t!]
  \centering
  \includegraphics[width=0.48\textwidth]{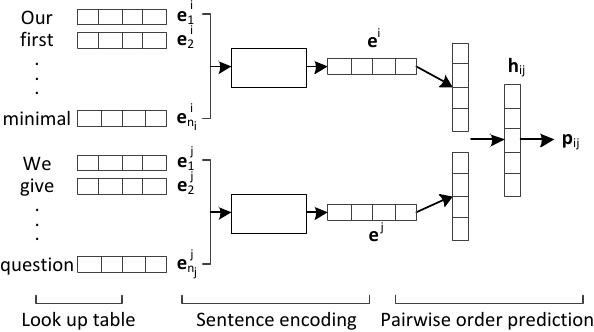}
  \caption{Neural network approach for pairwise order prediction.}\label{fig:pair_wise_model}
\end{figure}
Sentence ordering task can be viewed as a ranking problem. In this paper, we adopt the prevalent pairwise ranking model \cite{schapire1998learning,furnkranz2003pairwise,zheng2007regression}. That is, the goal is to predict the order of any two sentences pair $(s_i, s_j)$ as shown in Figure \ref{fig:pair_wise_model}. Formally, given representations of sentences $\e^1, \dots, \e^{n_s}$, embeddings from a sentence encoder (Section \ref{sec:SentenceEncoding}), we model the probability $\mathbf{p}_{ij}$ that sentence $s_i$ precedes $s_j$ as:
\begin{align}
\h_{ij} &= \phi({\bW_h}^\intercal (\e^i \oplus \e^j) + \bb_h),\\
\mathbf{p}_{ij} &= \sigma ({\bW_p}^\intercal \h_{ij} + \bb_p), \label{eq:p_pair}
\end{align}
where $\bW_h \in \R^{2d_s \times h}$, $\bb_h \in \R^h$, $\bW_p \in \R^{h}$, $\bb_p \in \R$ are trainable parameters. $\sigma(\cdot)$ is sigmoid function and $\phi(\cdot)$ is tanh function.

The score of sentences in order $o$ can be calculated as a log-likelihood maximization problem \cite{chen2013pairwise}:
\begin{gather}
    \text{score} (s,o,i,j) = \log \mathbf{p}_{o_io_j},\\
  \text{Score} (s,o)= \sum_{i = 1}^{n_s} \sum_{j = i + 1}^{n_s} \text{score} (s,o,i,j),
\end{gather}
where score$(s,o,i,j)$ indicates the score for sentence pairs $(o_i,o_j)$ and Score$(s,o)$ indicates the score of sentences $s$ in order $o$. Clearly, this can be seen as a two-layer neural network.

\begin{algorithm}[ht!]
\caption{Beam search for order prediction.}
\label{alg:beam_search}
\begin{algorithmic}[1]
\STATE beam = [ ]
\FOR{$i=1$; $i<=n_s$; $i++$}
    \STATE item = ([$i$], 0.0) \emph{\# tuple (partial order, award)}
    \STATE beam.append(item)
\ENDFOR
\STATE \emph{\# $n+1$ elements have been generated}
\FOR{$n=0$; $n<n_s-1$; $n++$}
    \STATE new\_beam = [ ]
        \FORALL {item $\in$ beam}
        \FOR{$j=1$; $j<=n_s$; $j++$}
            \STATE order = item[0] + [j] \emph{\# append new}
            \IF{any\_duplicate(order)==True}
                \STATE continue
            \ENDIF
            \STATE award = item[1]
            \FOR{$i=0$; $i<=n$; $i++$}
                \STATE award += score$(s, \text{order}, i,n+1)$
            \ENDFOR
            \STATE new\_beam.append((order, award))
        \ENDFOR
    \ENDFOR
    \STATE beam = N-Best(new\_beam)
\ENDFOR
\STATE $\hat{o}$, Score$(s, \hat{o})$ = Best(beam)
\RETURN $\hat{o}$
\end{algorithmic}
\end{algorithm}

\subsection{Order Prediction}
%

The order prediction phase aims to figure out the predicted sentence order $\hat{o}$ which maximizes $\text{Score} (s,o)$:
\begin{align}
  \hat{o} &= \argmax_o \text{Score} (s,o), \label{eq:decode}
\end{align}

Search all valid permutations by brute force to discover the optimal $\hat{o}$ is computationally expensive and fundamentally non-scalable. Therefore, we use the beam-search strategy to find a sub-optimal order. The details are show in Algorithm \ref{alg:beam_search}.
\section{Sentence Encoding} \label{sec:SentenceEncoding}
To figure out the impacts of various sentence representations, we employ three different sentence encoders to model sentences: continuous bag of words (CBoW), convolutional neural networks (CNN) and long short-term (LSTM) neural networks. All these models map words into a embedded space by looking up a embedding table.
\subsection{Continues Bag of Words}
Continues bag of words (CBoW) model \cite{mikolov2013efficient} simply averages the embeddings of words of a sentence. Formally, given the embeddings of $n_w$ words of a sentence $\e_1, \dots, \e_{n_w}$, we can get sentence embedding $\e$ by an average operation:
\begin{equation}
  \e = \frac{1}{n_w}\sum_{k=1}^{n_w} \e_k,
\end{equation}
where $\e \in \R^{d_s}$ and $\e_k \in \R^{d}$, and $d_s = d$ are dimensionalities of sentence embedding and word embeddings respectively.
\subsection{Convolutional Neural Networks}
Convolutional neural networks (CNNs) \cite{simard2003best} extract local features and gain the global prominent features by a max-pooling operation over sentence. Formally, we  represent sentence as:
\begin{align}
    \mathbf{cov}_k &= \phi (\bW_{cov}^\intercal (\oplus_{u = 0} ^ {l_f - 1} \e_{k + u}) + \bb_{cov}), \\
    \e &= \max_k \mathbf{cov}_k, \label{eq:max_pooling}
\end{align}
where $\bW_{cov} \in \R^{(d \times l_f) \times d_f}$ and $\bb_{cov} \in \R^{d_f}$ are trainable parameters, and $\phi(\cdot)$ is tanh function. Here, $k = 1, \dots, n_w - l_f + 1$, and $l_f$ and $d_f$ are hyper-parameters indicating the filter length and number of feature maps respectively. Notably, $\max$ operation in Eq (\ref{eq:max_pooling}) is a element-wise operation.

\subsection{Long Short-term Neural Networks}
Long short-term (LSTM) neural networks \cite{hochreiter1997long} aim to maintain the crucial information through time. LSTM is an advanced recurrent neural network (RNN), which alleviates the problem of gradient vanishment and explosion. Formally, LSTM has memory cells $\mathbf{c} \in \R^{d_r}$ controlled by three kinds of gates: input gate $\mathbf{i} \in \R^{d_r}$, forget gate $\mathbf{f} \in \R^{d_r}$ and output gate $\mathbf{o} \in \R^{d_r}$:
\begin{align}
\left[ \begin{array}{c}
            \mathbf{i}_t \\
            \mathbf{o}_t \\
            \mathbf{f}_t \\
            \tilde{\mathbf{c}}_t
            \end{array}\right] &= \left[\begin{array}{c}
                                            \sigma\\
                                            \sigma\\
                                            \sigma\\
                                            \phi
                                        \end{array}\right]
                                                 \left( {{\bW}_g}^\intercal
                                                                \left[\begin{array}{c}
                                                                        \e_t\\
                                                                        {\mathbf{h}}_{t-1}
                                                                \end{array}\right]
                                                        + {\bb}_g
                                                \right), \\
               \mathbf{c}_t    &= \mathbf{c}_{t - 1} \odot \mathbf{f}_t + \tilde{\mathbf{c}}_t \odot \mathbf{i}_t, \\
               {\mathbf{h}}_t &= \mathbf{o}_t \odot \phi( \mathbf{c}_t ),
\end{align}
where ${\bW}_g \in \R^{(d + d_r) \times 4d_r}$ and ${\bb}_g \in \R^{4d_r}$ are trainable parameters. $d_r$ is a hyper-parameter indicating the cell unit size as well as gate unit size.
$\sigma(\cdot)$ is sigmoid function and $\phi(\cdot)$ is tanh function.
Here, $t = 1, \dots, n_w$.
Thus, we would represent sentence as:
\begin{equation}
  \e = {\mathbf{h}}_{n_w}.
\end{equation}

\section{Training}
In this paper, we use pairwise ranking model. Thus, we extract $m$ gold sentence pairs \{$x^i$ = ($s^i_{fir}$, $s^i_{sec}$), $y^i$ = 1\}$_{i=1}^m$ as positive samples from the whole corpus. Meanwhile, we construct $m$ negative samples by reversing the gold sentence pairs \{$x^{i+m}$ = ($s^i_{sec}$, $s^i_{fir}$), $y^{i+m}$ = 0\}$_{i=1}^m$.

The objective is to minimize the loss function $J(\theta)$:
\begin{equation}
  J(\theta) = -\frac{1}{2m}\sum_{i=1}^{2m} y^i\log\p_{x^i} + (1-y^i)\log(1-\p_{x^i}), \label{eq:loss_func}
\end{equation}
where $\p_{x^i}$ is the probability that sentence pair $x^i$ is in correct order as Eq (\ref{eq:p_pair}). Here, parameter set $\theta$ indicates all trainable parameters of our model.

We use shuffled mini-batch stochastic gradient descent (SGD) algorithm together with adadelta \cite{zeiler2012adadelta} to train our model.

\begin{table}[t]\small\setlength{\tabcolsep}{1pt}
\centering
\begin{tabular}{|c|r|r|r|}
 \hline
   Attributes & Train &Dev& Test\\
 \hline
    \# of Abstracts &884,912&110,614&110,615\\
    \# of Sentences per Abstracts&5.38&5.39&5.37\\
\# of Words per Abstracts&134.58&134.80&134.58\\
 \hline
 \end{tabular}
\caption{Details of arXiv datasets.}\label{tab:info_dataset1}
\end{table}
\begin{table}[t]\small\setlength{\tabcolsep}{1pt}
\centering
\begin{tabular}{|c|r|r|r|}
 \hline
   Categories (Abbreviation) & Train & Dev & Test\\
 \hline
    Statistics (stat)&19,223&2,465&2,497\\
    Quantitative Biology (qbio)&15,495&1,943&1,866\\
Physics (phys)&821,795&102,584&102,892\\
Computer Science (cs)&84,689&10,624&10,453\\
Nonlinear Sciences (nlin)&13,273&1,619&1,695\\
Quantitative Finance (qfin)&5,201&708&670\\
Mathematics (math)&216,153&26,819&26,854\\
 \hline
 \end{tabular}
\caption{Details of category information of arXiv datasets. Notably, since categories could be overlapped, the size of total data set is smaller than the sum of numbers of all 7 categories.}\label{tab:info_dataset2}
\end{table}
\begin{table}[t] \setlength{\tabcolsep}{3pt}
\centering
\begin{tabular}{|c|c|}
  \hline
  Initial learning rate &$\alpha = 0.2$\\
  Hidden layer size&$h=100$\\
  Filter length of CNN&$l_f = 3$\\
    Batch size&$128$\\
  Beam size&128\\
  \hline
\end{tabular}
\caption{Hyper-parameter configurations.}\label{tab:paramSet}
\end{table}

\section{Experiments}
\subsection{Dataset}
Since abstracts of paper are always well written and have strong logic clues, we evaluate our models on all abstracts on arXiv website up to date\footnote{We collect all abstracts of paper before 2016-5-25.}. Abstracts from arXiv can be mainly classified into 7 categories: statistics, quantitative biology, physics, computer science, nonlinear sciences, quantitative finance and mathematics. The development set and test set are the first and last 10\% abstracts from shuffled data, and the training set consists of the remains. The detailed information of arXiv dataset is shown in Table \ref{tab:info_dataset1} and Table \ref{tab:info_dataset2}. We use NLTK toolkit \cite{bird2006nltk} to break paragraph into sentences.

\subsection{Hyper-parameters}
Table \ref{tab:paramSet} gives the details of hyper-parameter configurations.  Regularization term with coefficient $\lambda = 10^{-4}$ is omitted in Eq (\ref{eq:loss_func}) for simplicity. Besides, we set number of feature maps $d_f$ of CNN and cell unit size $d_r$ of LSTM as same as word embedding dimensionality $d$.

\begin{table*} \small \setlength{\tabcolsep}{3pt}
\centering
\begin{tabular}{|c|*{12}{c|}}
 \hline
 Metrics    &   \multicolumn{3}{c|}{Rouge-S}   &   \multicolumn{3}{c|}{Rouge-2}   &   \multicolumn{3}{c|}{Rouge-3}   &   \multicolumn{3}{c|}{P-all}\\
 \hline
 Models &CBoW&CNN&LSTM&CBoW&CNN&LSTM&CBoW&CNN&LSTM&CBoW&CNN&LSTM\\
 \hline
 25w &0.7993&0.8004&0.8217&0.4421&0.4416&0.4742&0.2420&0.2420&0.2729&0.2881&0.2888&0.3178\\
 50w &\textbf{0.8002}&0.8113&0.8278&\textbf{0.4438}&0.4579&0.4827&\textbf{0.2437}&0.2574&0.2818&\textbf{0.2892}&0.3022&0.3257\\
 100w&0.7982&0.8164&0.8296&0.4426&0.4669&0.4899&0.2423&0.2664&0.2892&0.2870&0.3114&0.3314\\
 200w&0.7992&\textbf{0.8192}&\textbf{0.8297}&0.4422&\textbf{0.4729}&\textbf{0.4916}&0.2420&\textbf{0.2716}&\textbf{0.2911}&0.2866&\textbf{0.3156}&\textbf{0.3343}\\
 \hline
 \hline
 Random&\multicolumn{3}{c|}{0.4999}   &   \multicolumn{3}{c|}{0.2309}   &   \multicolumn{3}{c|}{0.0582}   &   \multicolumn{3}{c|}{0.0807}\\
  \hline
 \end{tabular}
\caption{Performances of different models on test set of arXiv dataset.}\label{tab:res_all}
\end{table*}

\subsection{Evaluation Metrics}
To evaluation the results (predicted orders), we use three types of metrics: Rouge-S, Rouge-N \cite{lin2004rouge} and P-all. Unlike summarization task, the precision and recall rates are always the same in sentence ordering task. Thus, Rouge-S, Rouge-N could be introduced in a simpler way. Moreover, we also introduce P-all metric to calculate the ratio of exact matching orders.
\subsubsection{Rouge-S}
Rouge-S is skip-bigram co-occurrence statistics. Skip-bigram contains any pair of sentences in text, allowing for arbitrary gaps. Suppose we have a corpus including $M$ texts $s^1, s^2, \dots, s^M$. Then, Rouge-S could be formalized as:
\begin{equation}
  \text{Rouge-S} = \frac{1}{M} \sum_{m=1}^{M} \frac{|\text{S}(s^m, \hat{o}^m) \bigcap \text{S}(s^m, {o^m}^*)|}{|\text{S}(s^m, {o^m}^*)|},
\end{equation}
where S$(\cdot)$ is the set of all skip bigram sentence pairs of a text. Here, $s^m$ is the $m$-th text. $\hat{o}^m$ and ${o^m}^*$ are predicted and gold orders of $m$-th text respectively.
\subsubsection{Rouge-N}
Rouge-N is n-gram co-occurrence statistics which could be formalized as:
\begin{equation}
  \text{Rouge-N} = \frac{1}{M} \sum_{m=1}^{M} \frac{|\text{N}(s^m, \hat{o}^m) \bigcap \text{N}(s^m, {o^m}^*)|}{|\text{N}(s^m, {o^m}^*)|},
\end{equation}
where N$(\cdot)$ is the set of all N consecutive sentences in a given order.
\subsubsection{P-all}
P-all aims to calculate the radio of exact matching orders which could be formalized as:
\begin{equation}
  \text{P-all} = \frac{1}{M} \sum_{m=1}^{M} \textbf{1}\{\hat{o}^m = {o^m}^*\},
\end{equation}
where \textbf{1}$\{\cdot\}$ is indicator function.

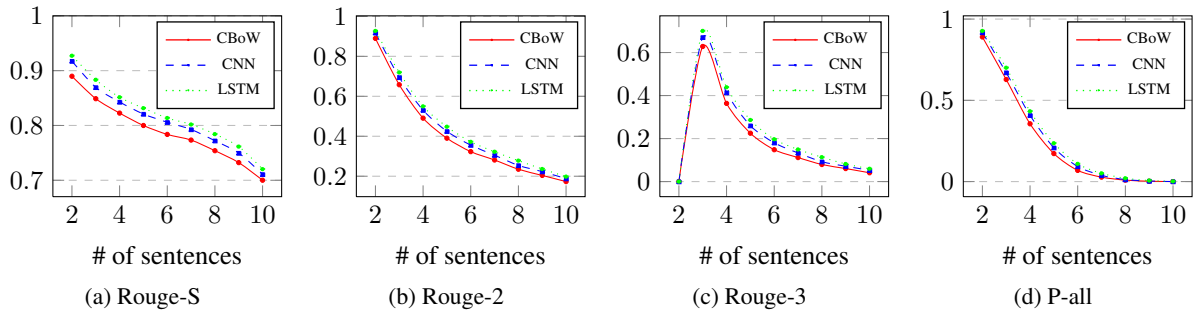
\begin{figure*} \small
  \centering
  \pgfplotsset{width=0.28\textwidth}
\subfloat[Rouge-S]{
  \begin{tikzpicture}
    \begin{axis}[
    xlabel={\# of sentences},
    ymax=1,
    legend entries={CBoW,CNN,LSTM,Random},
    mark size=0.8pt,
    ymajorgrids=true,
    grid style=dashed,
    legend pos= north east,
    legend style={font=\tiny,line width=.5pt,mark size=.2pt,
            /tikz/every even column/.append style={column sep=0.5em}},
            smooth,
    ]
    \addplot [red,mark=*] table [x index=0, y index=1] {Rouge-S_length_analyse.txt};
    \addplot [blue,dashed,mark=square*] table [x index=0, y index=2] {Rouge-S_length_analyse.txt};
    \addplot [green,dotted,mark=otimes*] table [x index=0, y index=3] {Rouge-S_length_analyse.txt};
    \end{axis}
\end{tikzpicture}
}
\hspace{0em}
\subfloat[Rouge-2]{
  \begin{tikzpicture}
    \begin{axis}[
    xlabel={\# of sentences},
    legend entries={CBoW,CNN,LSTM,Random},
    mark size=0.8pt,
    ymajorgrids=true,
    grid style=dashed,
    legend pos= north east,
    legend style={font=\tiny,line width=.5pt,mark size=.2pt,
            /tikz/every even column/.append style={column sep=0.5em}},
            smooth,
    ]
    \addplot [red,mark=*] table [x index=0, y index=1] {Rouge-2_length_analyse.txt};
    \addplot [blue,dashed,mark=square*] table [x index=0, y index=2] {Rouge-2_length_analyse.txt};
    \addplot [green,dotted,mark=otimes*] table [x index=0, y index=3] {Rouge-2_length_analyse.txt};
    \end{axis}
\end{tikzpicture}
}
\hspace{0em}
\subfloat[Rouge-3]{
  \begin{tikzpicture}
    \begin{axis}[
    xlabel={\# of sentences},
    legend entries={CBoW,CNN,LSTM,Random},
    mark size=0.8pt,
    ymajorgrids=true,
    grid style=dashed,
    legend pos= north east,
    legend style={font=\tiny,line width=.5pt,mark size=.2pt,
            /tikz/every even column/.append style={column sep=0.5em}},
            smooth,
    ]
    \addplot [red,mark=*] table [x index=0, y index=1] {Rouge-3_length_analyse.txt};
    \addplot [blue,dashed,mark=square*] table [x index=0, y index=2] {Rouge-3_length_analyse.txt};
    \addplot [green,dotted,mark=otimes*] table [x index=0, y index=3] {Rouge-3_length_analyse.txt};
    \end{axis}
\end{tikzpicture}\label{fig:rouge3_length}
}
\hspace{0em}
\subfloat[P-all]{
  \begin{tikzpicture}
    \begin{axis}[
    xlabel={\# of sentences},
    legend entries={CBoW,CNN,LSTM,Random},
    mark size=0.8pt,
    ymajorgrids=true,
    grid style=dashed,
    legend pos= north east,
    legend style={font=\tiny,line width=.5pt,mark size=.2pt,
            /tikz/every even column/.append style={column sep=0.5em}},
            smooth,
    ]
    \addplot [red,mark=*] table [x index=0, y index=1] {P-all_length_analyse.txt};
    \addplot [blue,dashed,mark=square*] table [x index=0, y index=2] {P-all_length_analyse.txt};
    \addplot [green,dotted,mark=otimes*] table [x index=0, y index=3] {P-all_length_analyse.txt};
    \end{axis}
\end{tikzpicture}
}
\caption{Performances of different sentence encoders with 200 dimensional word embeddings on different numbers of sentences on test set of arXiv dataset.}\label{fig:res_length}
\end{figure*}
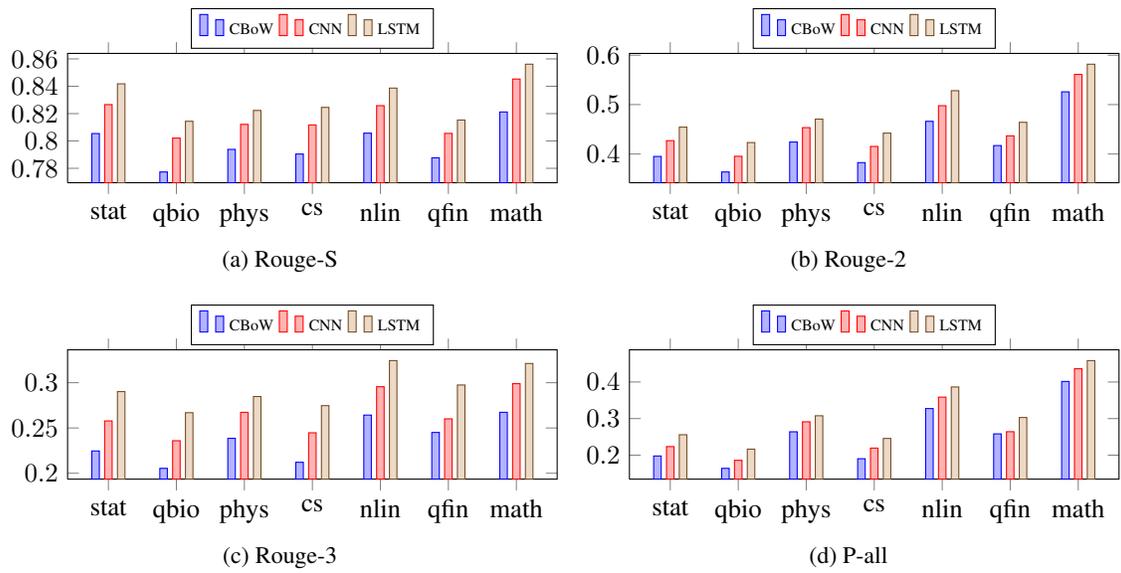
\begin{figure*}\small
\pgfplotstableread[row sep=\\,col sep=&]{
No	&	CBoW	&	CNN	&	LSTM	&	Random\\
stat	&	0.805424	&	0.826587	&	0.841777	&	0.504008\\
qfin	&	0.787683	&	0.805544	&	0.815293	&	0.513072\\
qbio	&	0.777343	&	0.802112	&	0.814396	&	0.498411\\
cs	&	0.790447	&	0.811700	&	0.824561	&	0.497799\\
nlin	&	0.805718	&	0.825828	&	0.838664	&	0.505307\\
phys	&	0.793839	&	0.812143	&	0.822316	&	0.499731\\
math	&	0.821212	&	0.845282	&	0.856138	&	0.500813\\
    }\RougeS
\pgfplotstableread[row sep=\\,col sep=&]{
No	&	CBoW	&	CNN	&	LSTM	&	Random\\
stat	&	0.395133	&	0.427019	&	0.454438	&	0.192170\\
qfin	&	0.417122	&	0.436848	&	0.464207	&	0.220966\\
qbio	&	0.363753	&	0.395453	&	0.423201	&	0.177537\\
cs	&	0.382622	&	0.415313	&	0.442365	&	0.187876\\
nlin	&	0.466165	&	0.497786	&	0.528137	&	0.247837\\
phys	&	0.424307	&	0.453404	&	0.470646	&	0.222386\\
math	&	0.525817	&	0.561069	&	0.581699	&	0.275523\\
    }\Rougetwo
\pgfplotstableread[row sep=\\,col sep=&]{
No	&	CBoW	&	CNN	&	LSTM	&	Random\\
stat	&	0.224630	&	0.257814	&	0.290186	&	0.045189\\
qfin	&	0.245072	&	0.260071	&	0.297482	&	0.060373\\
qbio	&	0.205419	&	0.235920	&	0.266871	&	0.038904\\
cs	&	0.212122	&	0.244711	&	0.274672	&	0.042807\\
nlin	&	0.264234	&	0.295561	&	0.324413	&	0.070895\\
phys	&	0.238586	&	0.267272	&	0.284722	&	0.058479\\
math	&	0.267316	&	0.298928	&	0.321231	&	0.064140\\
    }\Rougethree
\pgfplotstableread[row sep=\\,col sep=&]{
No	&	CBoW	&	CNN	&	LSTM	&	Random\\
stat	&	0.197837404886	&	0.223468161794	&	0.255907088506	&	0.0360432519023\\
qfin	&	0.258208955224	&	0.264179104478	&	0.302985074627	&	0.0671641791045\\
qbio	&	0.163987138264	&	0.185959271168	&	0.216505894962	&	0.0278670953912\\
cs	&	0.190280302306	&	0.219362862336	&	0.245766765522	&	0.0396058547785\\
nlin	&	0.327433628319	&	0.358702064897	&	0.386430678466	&	0.0979351032448\\
phys	&	0.263951828948	&	0.291242082071	&	0.307716381663	&	0.0695025162114\\
math	&	0.40150443137	&	0.436210620392	&	0.458516422135	&	0.133872048857\\
    }\Pall
\centering
\pgfplotsset{width=0.49\textwidth}
\pgfplotsset{height=.2\textwidth}
\subfloat[Rouge-S]{
\begin{tikzpicture}
    \begin{axis}[
            ybar,
            bar width=.1cm,
            legend style={font=\tiny, at={(0.5,1.35)},
                anchor=north,legend columns=-1},
            symbolic x coords={stat,qbio,phys,cs,nlin,qfin,math},
            xtick=data,
        ]
        \addplot table[x=No,y=CBoW]{\RougeS};
        \addplot table[x=No,y=CNN]{\RougeS};
        \addplot table[x=No,y=LSTM]{\RougeS};
        \legend{CBoW, CNN, LSTM}
    \end{axis}
\end{tikzpicture}
}
\hspace{0em}
\subfloat[Rouge-2]{
\begin{tikzpicture}
    \begin{axis}[
            ybar,
            bar width=.1cm,
            legend style={font=\tiny, at={(0.5,1.35)},
                anchor=north,legend columns=-1},
            symbolic x coords={stat,qbio,phys,cs,nlin,qfin,math},
            xtick=data,
        ]
        \addplot table[x=No,y=CBoW]{\Rougetwo};
        \addplot table[x=No,y=CNN]{\Rougetwo};
        \addplot table[x=No,y=LSTM]{\Rougetwo};
        \legend{CBoW, CNN, LSTM}
    \end{axis}
\end{tikzpicture}
}

\subfloat[Rouge-3]{
\begin{tikzpicture}
    \begin{axis}[
            ybar,
            bar width=.1cm,
            legend style={font=\tiny, at={(0.5,1.35)},
                anchor=north,legend columns=-1},
            symbolic x coords={stat,qbio,phys,cs,nlin,qfin,math},
            xtick=data,
        ]
        \addplot table[x=No,y=CBoW]{\Rougethree};
        \addplot table[x=No,y=CNN]{\Rougethree};
        \addplot table[x=No,y=LSTM]{\Rougethree};
        \legend{CBoW, CNN, LSTM}
    \end{axis}
\end{tikzpicture}
}
\hspace{0em}
\subfloat[P-all]{
\begin{tikzpicture}
    \begin{axis}[
            ybar,
            bar width=.1cm,
            legend style={font=\tiny, at={(0.5,1.35)},
                anchor=north,legend columns=-1},
            symbolic x coords={stat,qbio,phys,cs,nlin,qfin,math},
            xtick=data,
        ]
        \addplot table[x=No,y=CBoW]{\Pall};
        \addplot table[x=No,y=CNN]{\Pall};
        \addplot table[x=No,y=LSTM]{\Pall};
        \legend{CBoW, CNN, LSTM}
    \end{axis}
\end{tikzpicture}
}
\caption{Performances of different sentence encoders with 200 dimensional word embeddings on different categories on test set of arXiv dataset.}\label{fig:res_cate}
\end{figure*}
\begin{figure*}
  \centering
  \subfloat[CNN for sentence 1]{
  \includegraphics[width=0.23\textwidth]{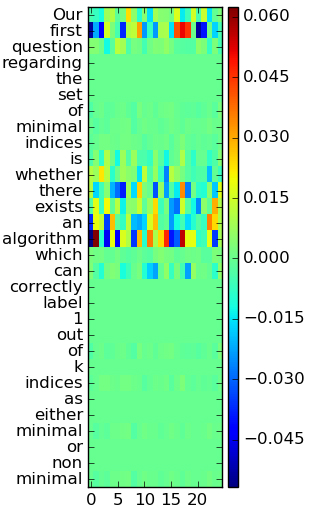} \label{fig:CNN1}
  }
  \hspace{0em}
  \subfloat[CNN for sentence 2]{
  \includegraphics[width=0.23\textwidth]{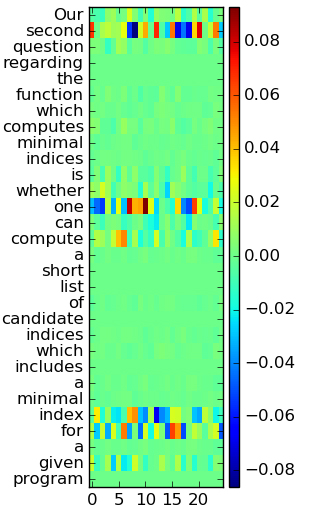} \label{fig:CNN2}
  }
  \subfloat[LSTM for sentence 1]{
  \includegraphics[width=0.23\textwidth]{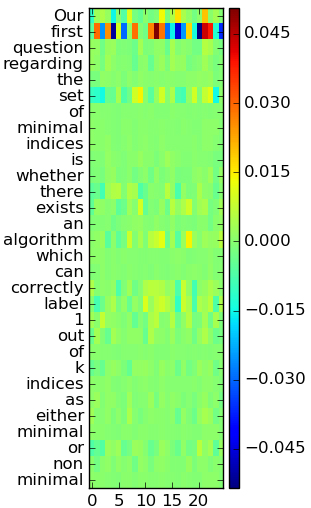} \label{fig:LSTM1}
  }
  \hspace{0em}
  \subfloat[LSTM for sentence 2]{
  \includegraphics[width=0.23\textwidth]{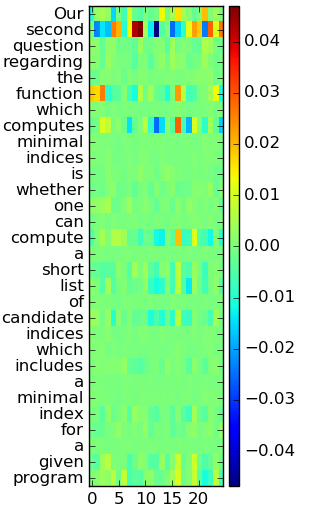} \label{fig:LSTM2}
  }
  \caption{Visualization of sentence 1 and sentence 2 using CNN and LSTM.}\label{fig:sentence_visual}
\end{figure*}

\begin{table*}\small\setlength{\tabcolsep}{3pt}
\centering
\begin{tabular}{|cc|cc|cc|}
\hline
\multicolumn{2}{|c|}{CBoW}&\multicolumn{2}{c|}{CNN}&\multicolumn{2}{c|}{LSTM}\\
\hline
2&\multicolumn{1}{p{0.28\textwidth}|}{
\textbf{
\color{white!55.00!blue}{Our} \color{white!32.42!blue}{second} \color{white!55.00!blue}{question} \color{white!55.00!blue}{regarding} \color{white!64.12!blue}{the} \color{white!36.22!blue}{function} \color{white!48.33!blue}{which} \color{white!36.22!blue}{computes} \color{white!72.85!blue}{minimal} \color{white!59.39!blue}{indices} \color{white!69.29!blue}{is} \color{white!66.74!blue}{whether} \color{white!32.42!blue}{one} \color{white!56.25!blue}{can} \color{white!32.42!blue}{compute} \color{white!46.15!blue}{a} \color{white!32.42!blue}{short} \color{white!32.42!blue}{list} \color{white!52.04!blue}{of} \color{white!32.42!blue}{candidate} \color{white!59.39!blue}{indices} \color{white!48.33!blue}{which} \color{white!32.42!blue}{includes} \color{white!46.15!blue}{a} \color{white!72.85!blue}{minimal} \color{white!42.22!blue}{index} \color{white!32.42!blue}{for} \color{white!46.15!blue}{a} \color{white!42.91!blue}{given} \color{white!32.42!blue}{program}
}
}
&1&\multicolumn{1}{p{0.28\textwidth}|}{
\textbf{
\color{white!19.17!blue}{Our} \color{white!24.56!blue}{first} \color{white!63.82!blue}{question} \color{white!80.00!blue}{regarding} \color{white!68.48!blue}{the} \color{white!80.00!blue}{set} \color{white!68.22!blue}{of} \color{white!47.59!blue}{minimal} \color{white!39.62!blue}{indices} \color{white!8.44!blue}{is} \color{white!26.63!blue}{whether} \color{white!38.82!blue}{there} \color{white!50.54!blue}{exists} \color{white!38.67!blue}{an} \color{white!6.67!blue}{algorithm} \color{white!46.74!blue}{which} \color{white!40.40!blue}{can} \color{white!80.00!blue}{correctly} \color{white!80.00!blue}{label} \color{white!80.00!blue}{1} \color{white!80.00!blue}{out} \color{white!68.22!blue}{of} \color{white!80.00!blue}{k} \color{white!39.62!blue}{indices} \color{white!69.27!blue}{as} \color{white!80.00!blue}{either} \color{white!47.59!blue}{minimal} \color{white!80.00!blue}{or} \color{white!80.00!blue}{non} \color{white!47.59!blue}{minimal}
}
}
&1&\multicolumn{1}{p{0.28\textwidth}|}{
\textbf{
\color{white!-0.00!blue}{Our} \color{white!18.93!blue}{first} \color{white!46.09!blue}{question} \color{white!64.92!blue}{regarding} \color{white!69.47!blue}{the} \color{white!57.09!blue}{set} \color{white!73.41!blue}{of} \color{white!69.50!blue}{minimal} \color{white!65.18!blue}{indices} \color{white!65.60!blue}{is} \color{white!56.08!blue}{whether} \color{white!69.37!blue}{there} \color{white!71.37!blue}{exists} \color{white!77.39!blue}{an} \color{white!64.53!blue}{algorithm} \color{white!77.68!blue}{which} \color{white!73.31!blue}{can} \color{white!68.54!blue}{correctly} \color{white!62.87!blue}{label} \color{white!69.97!blue}{1} \color{white!72.80!blue}{out} \color{white!73.41!blue}{of} \color{white!73.84!blue}{k} \color{white!65.18!blue}{indices} \color{white!75.78!blue}{as} \color{white!76.12!blue}{either} \color{white!69.50!blue}{minimal} \color{white!70.33!blue}{or} \color{white!77.63!blue}{non} \color{white!69.50!blue}{minimal}
}
}
\\
\hline
1&\multicolumn{1}{p{0.28\textwidth}|}{
\textbf{
\color{white!68.46!blue}{Our} \color{white!47.19!blue}{first} \color{white!68.46!blue}{question} \color{white!68.46!blue}{regarding} \color{white!74.37!blue}{the} \color{white!47.19!blue}{set} \color{white!71.45!blue}{of} \color{white!80.00!blue}{minimal} \color{white!71.93!blue}{indices} \color{white!79.04!blue}{is} \color{white!77.78!blue}{whether} \color{white!47.19!blue}{there} \color{white!47.19!blue}{exists} \color{white!47.76!blue}{an} \color{white!47.19!blue}{algorithm} \color{white!61.79!blue}{which} \color{white!69.02!blue}{can} \color{white!47.19!blue}{correctly} \color{white!47.19!blue}{label} \color{white!47.19!blue}{1} \color{white!47.19!blue}{out} \color{white!71.45!blue}{of} \color{white!47.19!blue}{k} \color{white!71.93!blue}{indices} \color{white!43.78!blue}{as} \color{white!47.19!blue}{either} \color{white!80.00!blue}{minimal} \color{white!55.95!blue}{or} \color{white!47.19!blue}{non} \color{white!80.00!blue}{minimal}
}
}
&3&\multicolumn{1}{p{0.28\textwidth}|}{
\textbf{
\color{white!7.69!blue}{We} \color{white!11.42!blue}{give} \color{white!34.79!blue}{some} \color{white!2.20!blue}{negative} \color{white!15.29!blue}{results} \color{white!39.34!blue}{and} \color{white!43.69!blue}{leave} \color{white!69.64!blue}{the} \color{white!78.79!blue}{possibility} \color{white!66.90!blue}{of} \color{white!47.08!blue}{positive} \color{white!15.29!blue}{results} \color{white!23.12!blue}{as} \color{white!22.47!blue}{open} \color{white!46.12!blue}{questions}
}
}
&2&\multicolumn{1}{p{0.28\textwidth}|}{
\textbf{
\color{white!33.73!blue}{Our} \color{white!34.32!blue}{second} \color{white!62.03!blue}{question} \color{white!69.66!blue}{regarding} \color{white!73.66!blue}{the} \color{white!63.01!blue}{function} \color{white!77.63!blue}{which} \color{white!60.21!blue}{computes} \color{white!72.92!blue}{minimal} \color{white!69.75!blue}{indices} \color{white!72.46!blue}{is} \color{white!68.64!blue}{whether} \color{white!77.73!blue}{one} \color{white!75.45!blue}{can} \color{white!66.64!blue}{compute} \color{white!80.00!blue}{a} \color{white!74.17!blue}{short} \color{white!71.79!blue}{list} \color{white!75.26!blue}{of} \color{white!72.28!blue}{candidate} \color{white!69.75!blue}{indices} \color{white!77.63!blue}{which} \color{white!77.68!blue}{includes} \color{white!80.00!blue}{a} \color{white!72.92!blue}{minimal} \color{white!73.42!blue}{index} \color{white!78.46!blue}{for} \color{white!80.00!blue}{a} \color{white!75.09!blue}{given} \color{white!71.30!blue}{program}
}
}
\\
\hline
3&\multicolumn{1}{p{0.28\textwidth}|}{
\textbf{
\color{white!6.83!blue}{We} \color{white!-0.00!blue}{give} \color{white!-0.00!blue}{some} \color{white!-0.00!blue}{negative} \color{white!-0.00!blue}{results} \color{white!-0.00!blue}{and} \color{white!-0.00!blue}{leave} \color{white!52.96!blue}{the} \color{white!-0.00!blue}{possibility} \color{white!56.45!blue}{of} \color{white!-0.00!blue}{positive} \color{white!-0.00!blue}{results} \color{white!22.72!blue}{as} \color{white!-0.00!blue}{open} \color{white!-0.00!blue}{questions}
}
}
&2&\multicolumn{1}{p{0.28\textwidth}|}{
\textbf{
\color{white!-0.00!blue}{Our} \color{white!3.90!blue}{second} \color{white!47.87!blue}{question} \color{white!80.00!blue}{regarding} \color{white!69.95!blue}{the} \color{white!75.92!blue}{function} \color{white!75.15!blue}{which} \color{white!64.31!blue}{computes} \color{white!42.89!blue}{minimal} \color{white!48.60!blue}{indices} \color{white!16.80!blue}{is} \color{white!31.55!blue}{whether} \color{white!12.88!blue}{one} \color{white!29.47!blue}{can} \color{white!49.77!blue}{compute} \color{white!67.78!blue}{a} \color{white!80.00!blue}{short} \color{white!80.00!blue}{list} \color{white!72.48!blue}{of} \color{white!80.00!blue}{candidate} \color{white!48.60!blue}{indices} \color{white!75.15!blue}{which} \color{white!80.00!blue}{includes} \color{white!67.78!blue}{a} \color{white!42.89!blue}{minimal} \color{white!25.60!blue}{index} \color{white!27.14!blue}{for} \color{white!67.78!blue}{a} \color{white!57.97!blue}{given} \color{white!80.00!blue}{program}
}
}
&3&\multicolumn{1}{p{0.28\textwidth}|}{
\textbf{
\color{white!51.30!blue}{We} \color{white!39.84!blue}{give} \color{white!59.32!blue}{some} \color{white!38.97!blue}{negative} \color{white!60.72!blue}{results} \color{white!57.69!blue}{and} \color{white!53.64!blue}{leave} \color{white!76.14!blue}{the} \color{white!56.44!blue}{possibility} \color{white!76.67!blue}{of} \color{white!54.49!blue}{positive} \color{white!60.72!blue}{results} \color{white!69.60!blue}{as} \color{white!58.17!blue}{open} \color{white!38.74!blue}{questions}
}
}
\\
\hline
\end{tabular}
\caption{Case Study. Color indicates importance of words in order prediction. The more important the words are, the darker the color is.}\label{tab:paragraph_visual}
\end{table*}

\subsection{Results}
We use Rouge-S, Rouge-2, Rouge-3 and P-all metrics to evaluate our model with different sentence encoders. We also vary dimensionality of word embeddings, as shown in Table \ref{tab:res_all}. Line ``random'' means we randomly generate the orders for texts.

According to the results, we find the performances of CNN and LSTM increase with larger word embedding size, whereas the performance of CBoW peaks at 50. Among 3 sentence encoders, LSTM outperforms others in any case, which is much more effective than random baseline. 
Especially, LSTM achieves 0.3343 on P-all metric, which means more than one third texts could be ranked correctly (exactly matched), whereas random baseline only achieves 0.0807 on P-all metric.
Rouge-S is much higher than other metrics, since any correct pair of sentences with arbitrary gaps contributes to Rouge-S score. In general, P-all is harder than Rouge-3, then Rouge-2 and Rouge-S. However, we find P-all scores are always higher than Rouge-3 scores here. The reason is that the texts with 2 sentences contribute to P-all score, and their Rouge-3 scores are always 0 as shown in Figure \ref{fig:rouge3_length}.

\paragraph{Detailed Results}
Figure \ref{fig:res_length} summarizes our performance on different text sizes, with the embedding dimension as 200.
The x-axis of each sub figure indicates number of sentences.
Results show that performances drop rapidly when texts scale up (number of sentences increases). Generally speaking, texts with more sentences are more difficult to rank correctly.
Specifically, on P-all metric, CBoW, CNN and LSTM could achieve 0.8898, 0.9174 and 0.9272 with 2 sentences respectively, whereas random baseline only makes it at 0.4977. However, the performance drops rapidly. LSTM only achieves 0.0015 on P-all with 10 sentences to rank.
Notably, Rouge-3 score of texts with 2 sentences is 0 (Figure \ref{fig:rouge3_length}), since there is no 3-grams in this case.

In addition, we investigate the performance on different categories as shown in Figure \ref{fig:res_cate}.
Interestingly, according to the category analysis results, we find that mathematics and nonlinear sciences are easier than other categories.
Specifically, LSTM could achieves 0.4585 on P-all metric, which means nearly one half math texts could be predicted exactly.

Moreover, we observed that the beginning and the ending sentences are easier to discern \cite{mostafazadeh2016corpus} as shown in Table \ref{tab:begin_end}. P$_{\text{Begin}}$ and P$_{\text{End}}$ indicate the ratio of correct beginning and ending cases respectively. P$_{\text{Mean}}$ indicates the ratio of correct positions. Notably, results on Table \ref{tab:begin_end} are based on models with 200 dimensional embeddings.

\begin{table}[t]\small
\centering
\begin{tabular}{|c|c|c|c|}
 \hline
   Models & P$_{\text{Begin}}$ & P$_{\text{End}}$ & P$_{\text{Mean}}$\\
 \hline
    CBoW    &0.7837&0.5762&0.5263\\
    CNN    &0.8294&0.6079&0.5585\\
    LSTM    &0.8485&0.6237&0.5760\\
 \hline
    Random  &0.2306&0.2316&0.2307\\
 \hline
 \end{tabular}
\caption{The performance of discerning the beginning and the ending sentences of proposed models on test set of arXiv dataset.}\label{tab:begin_end}
\end{table}

\subsection{Case Study}
To gain further insight, we pick the abstract of the paper ``On approximate decidability of minimal programs'' \cite{teutsch2015approximate} for case study. First, we visualize which the key words in abstract are important in order prediction. Then, we visualize the importance of words in scoring a given sentence pair. All visualizations are based on model using 25 dimensional word embeddings trained on computer science data, and the selected abstract is from test set of computer science category.
\subsubsection{Text Level Visualization}
We choose the last three sentences of the abstract for visualization, as shown in Table \ref{tab:paragraph_visual}. The texts in displayed orders are predicted by CBoW, CNN and LSTM respectively, and the sequence numbers in front of sentences indicate gold orders. Color indicates importance of words in order prediction. The more important the words are, the darker they are coloured.

How to calculate the importance of words (color)? Inspired by the back-propagation strategy \cite{erhan2009visualizing,simonyan2013deep,li2015visualizing}, 
which measures how much each input unit contributes to the final decision, we can approximate the importance of words by their first derivatives. Given a text $s_1, \dots, s_n$, the embedding of $k$-th word $w^i_k$ in $i$-th sentence $s_i$ is $\e^i_k$. Then, we define $A_{ij}(w^i_k)$ as the importance of word $w^i_k$ in predicting the order of sentence pair $(s_i, s_j)$:
\begin{equation}
  A_{ij} (w^i_k) = \frac{\partial\p_{ij}}{\partial\e^i_k},
\end{equation}
where $\p_{ij} \in \R$ is described in Eq (\ref{eq:p_pair}).

Thus, we could define the importance of a word $A(w^i_k)$ in whole text as:
\begin{equation}
  A(w^i_k) = \sum_{j = i+1}^{n}\p_{ij}|A_{ij} (w^i_k)|,
\end{equation}
where $|\cdot|$ is the norm of vector, and we use second order norm here.

\paragraph{Discussion}
According to the result, words such as ``first'' and ``second'' are indicative, as they imply logic clues. Also, since we only take the last three sentences of the abstract, it is quite reasonable that the word ``results'' appears in the last one or two sentences. We also find CBoW makes mistake in predicting the order of sentence pair (1, 2). Specifically, if score $\p_{2,1}^{\text{CBoW}}$ indicates the reward of placing sentence 2 in front of sentence 1, we could list the detailed score information of sentence pair (1, 2) and its reverse:
\begin{table}[h] \small \setlength{\tabcolsep}{5pt}
\centering
\begin{tabular}{|c|c|c|}
 \hline

 Models    &   $\p_{1, 2}$   &   $\p_{2, 1}$\\
 \hline
 CBoW   & 0.4911  &  0.6097 \\
 CNN    & 0.7083  &  0.3449 \\
 LSTM   & 0.8744  &  0.1110 \\
 \hline
 \end{tabular}
\caption{Detailed score information of sentence pair (1, 2) and its reverse.}\label{tab:res_score_sentence_pair}
\end{table}

As shown in Table \ref{tab:res_score_sentence_pair}, CBoW believes the sentence order (2, 1) gets higher score than the reverse. CNN and LSTM correctly predict the order, and LSTM does so predict with high confidence, with scores of orders (1, 2) and (2, 1) as 0.8744 and 0.1110, respectively.
\subsubsection{Sentence Level Visualization}
To visualize the importance of words in predicting order of sentence pair explicitly, we print the word information $A_{ij} (w^i_k)$ of sentence 1 and sentence 2 in Figure \ref{fig:sentence_visual}. Since CBoW only takes a simple average operation, word information $A_{ij} (w^i_k)$ in a sentence is the same. Thus, we only plot the results of CNN and LSTM.

\paragraph{Discussion}
As shown in Figure \ref{fig:sentence_visual}, both CNN and LSTM notice the key words ``first'' and ``seconde''. However, CNN also concentrates on other words like ``algorithm'', ``one'' which may not be useful in deciding the order. As the result in Table \ref{tab:res_score_sentence_pair}, LSTM is more confident than CNN to rank the sentence 1 in front of sentence 2. In another word, LSTM may clearly capture more important clues or logical information than CNN.

\section{Related Work}

A fundamental problem in text generation is information ordering, including word and sentence ordering. Comparing with word ordering \cite{tillmann2000word,zhang2012syntax,zhang2015discriminative,schmaltz2016word}, sentence ordering is still less studied. Existing works of sentence ordering focus to improve the external and downstream applications, such as multi-document summarization and discourse coherence \cite{van1985semantic,grosz1995centering,van1999semantic,elsner2007unified,barzilay2008modeling}. There is also a lack of intrinsic evaluation for sentence ordering.

\newcite{barzilay2002inferring} proposed two naive sentence ordering techniques,  such as majority ordering and chronological ordering, in the context of multi-document summarization.
\newcite{lapata2003probabilistic} proposed  a probabilistic model that assumes the probability of any given sentence is determined by its adjacent sentence and learns constraints on sentence order from a corpus of domain specific texts.
\newcite{okazaki2004improving} improved chronological ordering by resolving antecedent sentences of arranged sentences and combining topical segmentation.
\newcite{bollegala2010bottom} presented a bottom-up approach to arrange sentences extracted for multi-document summarization. To capture the association and order of two textual segments (e.g. sentences), they defined four criteria: chronology, topical-closeness, precedence, and succession. 

Unlike these existing works, we propose a data-driven method to learn the order of sentences. We use neural models to encode sentences and learn the pairwise orders. The text order can be further found by a beam search process.

\section{Conclusions}

Although sentence ordering is an important factor in natural language generation, it still lacks of intrinsic evaluation for sentence ordering task. To address this, this paper introduces a new large corpus for evaluation of sentence ordering task. The corpus is a collection of abstracts of academic papers. We use this corpus to evaluate a range of neural models. These neural models perform well for judging the order of sentence pair, but perform relatively poor on the whole abstract. Therefore, sentence ordering is still a challenging problem.  We hope that our corpus provides valuable training data and a testbed for sentence ordering task.

In the future, we would like to integrate other ranking models like list-wise model for sentence ordering task.

\bibliography{emnlp2016}
\bibliographystyle{emnlp2016}

\end{document}